\newcommand{\Real}{\mathbb R}

\newcommand{\Eq}[1]{\begin{equation}#1\end{equation}}
\newcommand{\Vc}[1]{\mbox{\boldmath$#1$}}

\newcommand{\vx}{\Vc{x}}
\newcommand{\vX}{\Vc{X}}

\newcommand{\vY}{\Vc{Y}}

\newcommand{\mX}{\mathcal{X}}

\newcommand{\mM}{\mathcal{M}}

\newcommand{\mI}{\mathcal{I}}
\newcommand{\mP}{\mathcal{P}}

\DeclareMathAlphabet{\mathsfsl}{OT1}{cmss}{m}{sl}

\documentclass[12pt,onecolumn,journal,draftcls]{IEEEtran}

\newtheorem{thm}{Theorem}[section]
\newtheorem{lemma}{Lemma}[section]
\newtheorem{cor}{ Corollary}[section]

\usepackage{pifont,calc,amsmath,amsfonts,amsbsy,amssymb,tabularx,float,times}
\usepackage{cite,url,graphicx,subfigure,epsfig,algorithm,algorithmic,multirow}

\begin{document}

\title{An Information Geometric Framework for Dimensionality Reduction}%
\author{Kevin M. Carter$^1$, Raviv Raich$^2$, and Alfred O. Hero III$^1$\\
$^{1}$ Dept.~of EECS,    University of Michigan,    Ann Arbor,
MI 48109\\
$^{2}$ School of EECS, Oregon State University, Corvallis, OR 97331\\
{\normalsize \tt  \{kmcarter,hero\}@umich.edu}, {\normalsize \tt
raich@eecs.oregonstate.edu}}
\maketitle

\begin{abstract}
This report concerns the problem of dimensionality reduction through information geometric methods on statistical manifolds. While there has been considerable work recently presented regarding dimensionality reduction for the purposes of learning tasks such as classification, clustering, and visualization, these methods have focused primarily on Riemannian manifolds in Euclidean space. While sufficient for many applications, there are many high-dimensional signals which have no straightforward and meaningful Euclidean representation. In these cases, signals may be more appropriately represented as a realization of some distribution lying on a \emph{statistical} manifold, or a manifold of probability density functions (PDFs). We present a framework for dimensionality reduction that uses information geometry for both statistical manifold reconstruction as well as dimensionality reduction in the data domain.
\end{abstract}

\section{Introduction}
In the recent past, sensing and media storage capabilities have enabled the generation of enormous amounts of information, often in the form of high-dimensional data. This is easily viewed within sensor networks, imaging, and biomedical applications such as flow cytometry and gene micro-arrays. While this vast amount of retrieved data has opened a wealth of opportunities for data analysis, the problem of the \emph{curse of dimensionality} has become more substantial. The high dimensional nature of data is often simply a product of its representation. In many instances data dimensions are redundant and entirely correlated with some combination of other dimensions within the same data set. In these instances, although the retrieved data seems to exhibit a naturally high dimension, it is actually constrained to a lower dimensional subset -- manifold -- of the measurement space. This allows for significant dimension reduction with minor or no loss of information.

Often data does not exhibit a low intrinsic dimension in the data domain as one would have in manifold learning in Euclidean space. A straightforward strategy is to express the data in terms of a low-dimensional feature vector to alleviate the dimensionality issue. This initial processing of data as real-valued feature vectors in Euclidean space, which is often carried out in an ad hoc manner, has been called the ``dirty laundry'' of machine learning \cite{Dietterich:2002}. This procedure is highly dependent on having a good model for the data, and in the absence of such model may be highly suboptimal.

In this report we discuss an information geometric framework for dimensionality reduction. We view high-dimensional data sets as realizations of some generative model, or probability density function (PDF). Rather than dealing with Riemannian manifolds in a Euclidean space, we focus our attention on statistical manifolds for our geometric constructs. We offer two forms of dimensionality reduction: one which embeds these high-dimensional data sets into a single low-dimensional representation in Euclidean space by reconstructing the statistical manifold. This is performed by multidimensional scaling using information geometric measures of distance between PDFs. Secondly, we offer dimensionality reduction in the data domain by preserving the high-dimensionality similarities between data PDFs in the low-dimensional subspace. This is useful for both visualization and variable selection of high-dimensional data.

\section{Background on Information Geometry}
\label{S:Background}
Information geometry is a field that has emerged from the study of geometrical constructs on manifolds of probability distributions. These investigations analyze probability distributions as geometrical structures in a Riemannian space. Using tools and methods deriving from differential geometry, information geometry is applicable to information theory, probability theory, and statistics. The field of information theory is largely based on the works of Shun'ichi Amari \cite{Amari:90} and has been used for analysis in such fields as statistical inference, neural networks, and control systems. In this section, we will give a brief background on the methods of information geometry utilized throughout the rest of this report. For a more thorough introduction to information geometry, we suggest \cite{Kass&Vos:97,Amari&Nagaoka:2000}.
\subsection{Differential Manifolds}
\label{SS:DiffMan}
The concept of a differential manifold is similar to that of a smooth curve or surface lying in a high-dimensional space. A manifold $\mM$ can be intuitively thought of as a set of points with a coordinate system. These points can be from a variety of constructs, such as Euclidean coordinates, linear system, images, or probability distributions. Regardless of the definition of the points in the manifold $\mM$, there exists a coordinate system with a one-to-one mapping from $\mM$ to $\Real^d$, and as such, $d$ is known as the dimension of $\mM$.

For reference, we will refer to the coordinate system on $\mM$ as $\psi:\mM\rightarrow \Real^d$. If $\psi$ has $\mM$ as its domain, we call it a global coordinate system \cite{Amari&Nagaoka:2000}. In this situation, $\psi$ is a one-to-one mapping onto $\Real^d$ for all points in $\mM$. A manifold is differentiable if the coordinate system mapping $\psi$ is differentiable over its entire domain. If $\psi$ is infinitely differentiable, the manifold is said to be `smooth' \cite{Kass&Vos:97}.


In many cases there does not exist a global coordinate system. Examples of such manifolds include the surface of a sphere, the ``swiss roll'', and the torus. For these manifolds, there are only local coordinate systems. Intuitively, a local coordinate system acts as a global coordinate system for a local neighborhood of the manifold, and there may be many local coordinate systems for a particular manifold. Fortunately, since a local coordinate system contains the same properties as a global coordinate system (only on a local level), analysis is consistent between the two. As such, we shall focus solely on manifolds with a global coordinate system.
\subsubsection{Statistical Manifolds}
\label{SS:StatMan}
Let us now present the notion of statistical manifolds, or a set $\mM$ whose elements are probability distributions. A probability density function (PDF) on a set $\mX$ is defined as a function $p:\mX\rightarrow\Real$ in which
\[
p(x)\ge0, \,\forall x\in\mX
\]
\Eq{ \label{Equation:PDF}
\int p(x)\,dx=1
.}
We describe only the case for continuum on the set $\mX$, however if $\mX$ was discrete valued, equation (\ref{Equation:PDF}) will still apply by switching $\int p(x)\,dx=1$ with $\sum p(x)=1$. If we consider $\mM$ to be a family of PDFs on the set $\mX$, in which each element of $\mM$ is a PDF which can be parameterized by $\theta=\left[\theta^1,\ldots,\theta^n\right]$, then $\mM$ is known as a statistical model on $\mX$. Specifically, let
\Eq{ \label{E:statman}
\mM=\{p(x\mid\theta)\mid\theta\in\Theta\subseteq\Real^d\}
,}
with $p(x\mid\theta)$ satisfying the equations in (\ref{Equation:PDF}). Additionally, there exists a one-to-one mapping between $\theta$ and $p(x\mid\theta)$.

Given certain properties of the parameterization of $\mM$, such as differentiability and $C^\infty$ diffeomorphism (details of which are described in \cite{Amari&Nagaoka:2000}), the parameterization $\theta$ is also a coordinate system of $\mM$. In this case, $\mM$ is known as a statistical manifold. In the rest of this report, we will use the terms `manifold' and `statistical manifold' interchangeably.

\subsection{Fisher Information Distance}
\label{SS:FIM}
The Fisher information metric measures the amount of information a random variable $X$ contains in reference to an unknown parameter $\theta$. For the single parameter case it is defined as
\[
\mI(\theta)=E\left[\left(\frac{\partial}{\partial\theta}\log f(X;\theta)\right)^2|\theta\right]
.\]
If the condition $\int\frac{\partial^2}{\partial\theta^2}f(X;\theta)\,dX=0$ is met, then the above equation can be written as
\[
\mI(\theta)=-E\left[\frac{\partial^2}{\partial\theta^2}\log f(X;\theta)\right]
.\]
For the case of multiple parameters $\theta=\left[\theta^1,\ldots,\theta^n\right]$, we define the Fisher information matrix $[\mI(\theta)]$, whose elements consist of the Fisher information with respect to specified parameters, as
\Eq{ \label{Equation:FIM}
\left[\mI(\theta)\right]_{ij}=\int{f(X;\theta)\frac{\partial \log f(X;\theta)}{\partial\theta^i}\frac{\partial \log f(X;\theta)}{\partial\theta^j}\,dX}
.}

For a parametric family of probability distributions, it is possible to define a Riemannian metric using the Fisher information matrix, known as the information metric. The information metric distance, or Fisher information distance, between two distributions $p(x;\theta_1)$ and $p(x;\theta_2)$ in a single parameter family is
\Eq{ \label{Eq:FID_single}
    D_F(\theta_1,\theta_2)=\int_{\theta_1}^{\theta_2}{\mI(\theta)^{1/2}d\theta}
    ,}
where $\theta_1$ and $\theta_2$ are parameter values corresponding to the two PDFs and $\mI(\theta)$ is the Fisher information for the parameter $\theta$. Extending to the multi-parameter case, we obtain:
\Eq{ \label{Eq:FID_multi}
    D_F(\theta_1,\theta_2) =
\mathop{\mathop{\min_{\theta(\cdot):}}_{\theta(0)=\theta_1}}_{\theta(1)=\theta_2}
\int_0^1 \sqrt{\bigl(\frac{d\theta}{dt}\bigr)^T \left[{\cal I}(\theta)\right] \bigl(\frac{d\theta}{dt}\bigr)}\, dt
    .}
where $\theta=\theta(t)$ is the parameter path along the manifold. Note that the coordinate system of a statistical manifold is the same as the parameterization of the PDFs (i.e.~$\theta$). Essentially, (\ref{Eq:FID_multi}) amounts to finding the length of the shortest path -- the geodesic -- on $\mM$ connecting coordinates $\theta_1$ and $\theta_2$.

\subsection{Approximation of Fisher Information Distance}
\label{SS:ApproxFish}
The Fisher information distance is a consistent metric, regardless of the parameterization of the manifold \cite{Srivastava:CVPR07}. This fact enables the approximation of the information distance when the specific parameterization of the manifold is unknown, and there have been many metrics developed for this approximation. An important class of such divergences is known as the $f$-divergence \cite{Csiszar:67}, in which $f(u)$ is a convex function on $u > 0$ and
\[
    D_f(p\|q)=\int{p(x)f\left(\frac{q(x)}{p(x)}\right)dx}
    .\]

A specific and important example of the $f$-divergence is the $\alpha$-divergence, where $D^{(\alpha)}=D_{f^{(\alpha)}}$ for a real number $\alpha$. The function $f^{(\alpha)}(u)$ is defined as
\[
        f^{(\alpha)}(u)=\left\{
                          \begin{array}{cl}
                            \frac{4}{1-\alpha^2}\left(1-u^{(1+\alpha)/2}\right) & \alpha\neq\pm1 \\
                            u\log u & \alpha=1 \\
                            -\log u & \alpha=-1
                          \end{array}
                        \right.
.\]

As such, the $\alpha$-divergence can be evaluated as
\[
    D^{(\alpha)}(p\|q)=\frac{4}{1-\alpha^2}\left(1-\int{p(x)^\frac{1-\alpha}{2}q(x)^\frac{1+\alpha}{2}dx}\right)\quad\alpha\neq \pm1
,\]
and
\Eq{ \label{E:Dalpha-1}
    D^{(-1)}(p\|q)=D^{(1)}(q\|p)=\int{p(x)\log \frac{p(x)}{q(x)}dx}
.}
The $\alpha$-divergence is the basis for many important and well known divergence metrics, such as the Kullback-Leibler divergence and the Hellinger distance.
%

\subsubsection{Kullback-Leibler Divergence}
\label{SS:KLD}
The Kullback-Leibler (KL) divergence is defined as
\Eq{\label{Equation:KL}
KL(p\|q)=\int{p(x)\log \frac{p(x)}{q(x)}dx}
 ,}
which is equal to $D^{(-1)}$ (\ref{E:Dalpha-1}). The KL-divergence is a very important metric in information theory, and is commonly referred to as the relative entropy of one PDF to another. Kass and Vos show \cite{Kass&Vos:97} the relation between the Kullback-Leibler divergence and the Fisher information distance is
\[\sqrt{2KL(p\|q)}\rightarrow D_{F}(p,q)\]
as $p\rightarrow q$. This allows for an approximation of the Fisher information distance, through the use of the available PDFs, without the need for the specific parameterization of the manifold.

It should be noted that the KL-divergence is not a distance metric, as it does not satisfy the symmetry, $KL(p\|q)\neq KL(p\|q)$, or triangle inequality properties of a distance metric. To obtain symmetry, we will define the symmetric KL-divergence as:
\Eq{ \label{Equation:DKL}
        D_{KL}(p,q) = KL(p\|q)+KL(q\|p)
    ,}
which is symmetric, but still not a distance as it does not satisfy the triangle inequality. Since the Fisher information is a symmetric measure, we can relate the symmetric KL-divergence
and approximate the Fisher information distance as
\Eq{ \label{E:FisherApprox}
\sqrt{D_{KL}(p,q)}\rightarrow D_F(p,q)
,}
as $p\rightarrow q$.

\subsubsection{Hellinger Distance}
\label{SS:Hellinger}
Another important result of the $\alpha$-divergence is the evaluation with $\alpha=0$:
\[
    D^{(0)}(p\|q)=2\int{\left(\sqrt{p(x)}-\sqrt{q(x)}\right)^2dx},
\]
which is called the closely related to the Hellinger distance,
\[
    D_H=\sqrt{\frac{1}{2}D^{(0)}}
,\]
which satisfies the axioms of distance -- symmetry and the triangle inequality.  The Hellinger distance is related to the information distance in the limit by
\[2D_{H}(p,q)\rightarrow D_{F}(p,q)
\]
as $p\rightarrow q$ \cite{Kass&Vos:97}. We note that the Hellinger distance is related to the Kullback-Leibler divergence, as in the limit $\sqrt{KL(p\|q)}\rightarrow D_H(p,q)$.

\subsubsection{Other Fisher Approximations}
\label{SS:Others}
There are other metrics which approximate the Fisher information distance, such as the cosine distance. When dealing with multinomial distributions, the approximation
\[D_C(p,q)=2\arccos \int{\sqrt{p(x)\cdot q(x)}\,dx} \to D_F(p,q),\]
is the natural metric on the sphere.

Throughout this report we restrict our analysis to that of the Kullback-Leibler divergence and the Hellinger distance. The KL-divergence is a great means of differentiating shapes of continuous PDFs. Analysis of (\ref{Equation:KL}) shows that as $p(x)/q(x)\rightarrow \infty$, $KL(p\|q)\rightarrow \infty$. These properties ensure that the KL-divergence will be amplified in regions where there is a significant difference in the probability distributions. This cannot be used in the case of a multinomial PDF, however, because of divide-by-zero issues. In that case the Hellinger distance is the desired metric as there exists a monotonic transformation function $\psi:D_H\to D_C$ \cite{Kass&Vos:97}. Specific details on how we nonparametrically calculate these information divergences is provided in Section \ref{S:Details}. For additional measures of probabilistic distance, some of which approximate the Fisher information distance, and a means of calculating them between data sets, we refer the reader to \cite{Zhou&Chellappa:PAMN06}.

\section{Fisher Information Nonparametric Embedding}
\label{S:FINE}
Many applications of statistical manifolds have proved promising, such as document classification \cite{Lebanon:ISIGA05,Lafferty&Lebanon:JMLR05,Carter&Raich:ICASSP08}, flow cytometry analysis \cite{Finn&Carter:CytB08,Carter:MLSP08}, face recognition \cite{Arandjelovic:CVPR05}, texture segmentation \cite{Lee:ICIP05}, image analysis \cite{Srivastava:CVPR07}, clustering \cite{Salojarvi:ICANN03}, and shape analysis \cite{Kim:LIDS05}. While all have proposed alternatives to using Euclidean geometry for data modeling, most methods (outside of our own work) focus on clustering and classification, and do not explicitly address the problems of dimensionality reduction and visualization. Additionally, most presented work has been in the parametric setting, in which parameter estimation is a necessity for the various methods. This becomes ad-hoc and potentially troublesome if a good model is unavailable.

We provide a method of dimensionality reduction -- deemed \emph{Fisher Information Nonparametric Embedding} (FINE) -- which approaches the problem of statistical manifold reconstruction. This method includes a characterization of data sets in terms of a nonparametric statistical model, a geodesic approximation of the Fisher information distance as a metric for evaluating similarities between data sets, and a dimensionality reduction procedure to obtain a low-dimensional Euclidean embedding of the original high-dimensional data set for the purposes of both classification and visualization. This non-linear embedding is driven by information, not Euclidean, geometry. Our methods require no explicit model assumptions; only that the given data is a realization from an unknown model with some natural parameterization.
\subsection{Approximation of Distance on Statistical Manifolds}
\label{SS:ApproxDist}


Let us consider the approximation function $\hat{D}_F(p_1,p_2)$ of the Fisher information distance between $p_1$ and $p_2$, which may can be calculated using a variety of metrics as $p_1\to p_2$ (see Section \ref{SS:ApproxFish}). If $p_1$ and $p_2$ do not lie closely together on the manifold, these approximations become weak.
A good approximation can still be achieved if the manifold is densely sampled between the two end points. By defining the path between $p_1$ and $p_2$ as a series of connected segments and summing the length of those segments, we may approximate the length of the geodesic with graphical methods. Specifically, given the set of $n$ PDFs parameterized by $\mP_\theta=\left\{\theta_1,\ldots,\theta_N\right\}$, the Fisher information distance between $p_1$ and $p_2$ can be estimated as:
\[
D_F(p_1,p_2)\approx\min_{M,\left\{\theta_{(1)},\ldots,\theta_{(M)}\right\}} {\sum_{i=1}^{M-1}{ D_F(p(\theta_{(i)}),p(\theta_{(i+1)}))}},\quad p(\theta_{(i)})\rightarrow p(\theta_{(i+1)})\: \forall \: i
\]
where $p(\theta_{(1)})=p_1$, $p(\theta_{(M)})=p_2$, $\left\{\theta_{(1)},\ldots,\theta_{(M)} \right\} \in \mP_\theta$, and $M\leq N$.


Using an approximation of the Fisher information distance as $p_1\to p_2$, we can now define an approximation function $G$ for all pairs of PDFs:
\Eq{ \label{Eq:FID_Approx_G}
G(p_1,p_2;\mP)=\min_{M,\mP} {\sum_{i=1}^{M-1}{ \hat{D}_F(p_{(i)},p_{(i+1)})}}, \quad p_{(i)}\rightarrow p_{(i+1)}\: \forall \: i}
where $\mP=\left\{p_1,\ldots,p_N\right\}$ is the available collection of PDFs on the manifold. Intuitively, this estimate calculates the length of the shortest path between points in a connected graph on the well sampled manifold, and as such $G(p_1,p_2;\mP)\to D_F(p_1,p_2)$ as $N\to\infty$. This is similar to the manner in which Isomap \cite{Tenenbaum&etal:Science00} approximates distances on Riemannian manifolds in Euclidean space. 

\subsection{Dimensionality Reduction}
\label{SS:DimRed}
Given a matrix of dissimilarities between entities, many algorithms have been developed to find a low-dimensional embedding of the original data $\psi:\mM\rightarrow \Real^d$. These techniques have been classified as a group of methods called multidimensional scaling (MDS) \cite{Cox&Cox:94}. There are supervised methods \cite{Friedman:JASA89,Mika:NN99,Raich&Hero:ICASSP06,Goldberg&Roweis:NIPS04} which are generally used for classification purposes, and unsupervised methods \cite{Roweis&Saul:Science00,Belkin&Niyogi:NIPS02}, which are often used for clustering and manifold learning. In conjunction with the matrix of Fisher information distance approximations, these MDS methods allows us to find a single low-dimensional coordinate representation of each high-dimensional, large sample, data set. We do not highlight the heavily utilized Isomap \cite{Tenenbaum&etal:Science00} algorithm since it is identical to using classical MDS on the approximation of the geodesic distances.

\subsection{FINE Algorithm}
\label{SS:FINEAlgo}
We have identified a series of methods for manifold learning developed in the field of information geometry. By performing dimensionality reduction on a family of data sets, we are able to both better visualize and classify the data. In order to obtain a lower dimensional embedding, we calculate a dissimilarity metric between data sets within the family by approximating the Fisher information distance between their corresponding PDFs.

In problems of practical interest, however, the parameterization of the probability densities is usually unknown. We instead are given a family of data sets $\mX=\{\vX_1,\ldots,\vX_N\}$, in which we may assume that each data set $\vX_i$ is a realization of some underlying probability distribution to which we do not have knowledge of the parameters. As such, we rely on nonparametric techniques to estimate both the probability density and the approximation of the Fisher information distance. In the work presented in this report, we implement kernel density estimation methods (see Appendix \ref{A:KDE}), although $k$-NN methods are also applicable. Following these approximations, we are able to perform the same multidimensional scaling operations as previously described.

Fisher Information Nonparametric Embedding (FINE) is presented in Algorithm~\ref{algo:full} and combines all of the presented methods in order to find a low-dimensional embedding of a collection of data sets. If we assume each data set is a realization of an underlying PDF, and each of those distributions lie on a manifold with some natural parameterization, then this embedding can be viewed as an embedding of the actual manifold into Euclidean space. Note that in line \ref{a:embed}, `mds$(G,d)$' refers to using any multidimensional scaling method to embed the dissimilarity matrix $G$ into a Euclidean space with dimension $d$.

\begin{algorithm}[t]
\caption{Fisher Information Nonparametric Embedding}
\label{algo:full}
    \begin{algorithmic}[1]
        \REQUIRE Collection of data sets $\mX=\{\vX_1,\ldots,\vX_N\}$; the desired embedding dimension $d$
        \FOR{$i=1$ to $N$}
            \STATE Calculate $\hat{p}_i(\vx)$, the density estimate of $\vX_i$
        \ENDFOR
        \STATE Calculate $G$, where $G(i,j)$ is the geodesic approximation of the Fisher information distance between $p_i$ and $p_j$ (\ref{Eq:FID_Approx_G})
        \STATE $\vY=\textrm{mds}(G,d)$ \label{a:embed}
        \ENSURE $d$-dimensional embedding of $\mX$, into Euclidean space $\vY\in \Real^{d\times N}$
    \end{algorithmic}
\end{algorithm}

At this point it is worth stressing the benefits of this framework. Through information geometry, FINE enables the joint embedding of multiple data sets $\vX_i$ into a single low-dimensional Euclidean space. By viewing each $\vX_i\in\mX$ as a realization of $p_i\in\mP$, we reduce the numerous samples in $\vX_i$ to a single point. The dimensionality of the statistical manifold may be significantly less than that of the Euclidean realizations. For example, a Gaussian distribution is entirely defined by its mean $\mu$ and covariance $\Sigma$, leading to a 2-dimensional statistical manifold, while the dimensionality of the realization $\vX\sim N(\mu,\Sigma)$ may be significantly larger (i.e.$~\mu\in\Real^d$, $d\gg2$). MDS methods reduce the dimensionality of $p_i$ from the Euclidean dimension to the dimension of the statistical manifold on which it lies. This results in a single low-dimensional representation of each original data set $\vX_i\in\mX$.

\section{Information Preserving Component Analysis}
\label{S:IPCA}
We now extend our framework to dimensionality reduction in the data domain, which is the standard setting for manifold learning. However, rather than focusing on the relationships between elements in a single data set $\vX$, it is often desirable to compare each set in a collection $\mX=\left\{\vX_1,\ldots,\vX_N\right\}$ in which $\vX_i$ has $n_i$ elements $x\in\Real^d$. Once again, we wish to find a form of dimensionality reduction that preserves the information geometry of the statistical manifold of PDFs generating $\mX$. Unlike FINE, however, we now are interested in reduction of each $\vX_i$ individually to a low-dimensional subspace which preserves the Fisher information distances between $p_i$ and $p_j$, the estimated PDFs of $\vX_i$ and $\vX_j$ respectively. With an abuse of notation, we will further refer to $D_F(p_i,p_j)$ as $D_F(\vX_i,\vX_j)$ with the knowledge that the Fisher information distance is calculated with respect to PDFs, not realizations.

For this task, we are interested in linear and unsupervised methods, as class labels for $\mX$ are often unavailable and linear projections will be useful for variable selection. We define the \emph{Information Preserving Component Analysis (IPCA)} projection matrix $A\in\Real^{m\times d}$, in which $A$ reduces the dimension of $\vX$ from $d\rightarrow m$ ($m\leq d$), such that
\Eq{\label{e:IPCA_goal}
D_F(A\vX_i,A\vX_j)=D_F(\vX_i,\vX_j),\: \forall \: i,j
.}
Formatting as an optimization problem, we would like to solve:
\Eq{\label{E:proj_obj1}
A=\arg\min_{A:AA^T=I} J(A)
,}
where $I$ is the identity matrix and $J(A)$ is some cost function designed to implement (\ref{e:IPCA_goal}). Note that we include the optimization constraint $AA^T=I$ to ensure our projection is orthonormal, which keeps the data from scaling or skewing as that would undesirably distort the data. Let $D(\mX)$ be a dissimilarity matrix such that $D_{ij}(\mX)=D_F(\vX_i,\vX_j)$, and $D(\mX;A)$ is a similar matrix where the elements are perturbed by $A$, i.e.~$D_{ij}(\mX;A)=D_F(A\vX_i,A\vX_j)$. We have formulated several different cost functions with differing benefits:
\begin{eqnarray}
J_1(A) & = & \| D(\mX)-D(\mX;A)\|_F^2 \label{e:J1} \\
J_2(A) & = & \left\|\, {\rm e}^{-D(\mX)/c}-{\rm e}^{-D(\mX;A)/c}\right\|_F^2 \label{e:J2} \\
J_3(A) & = & -\| D(\mX;A) \|_F^2 \label{e:J3} \\
J_4(A) & = & \|\, {\rm e}^{-D(\mX;A)/c}\|_F^2 \label{e:J4}
,
\end{eqnarray}
where $\|\cdot\|_F$ is the Frobenius norm and $c$ is some constant.

It should be clear that $J_1(A)$ (\ref{e:J1}) is a direct implementation of our stated objective. $J_2(A)$ (\ref{e:J2}) applies a sense of locality to our objective, which is useful give that the approximations to the Fisher information distance are valid only as $p\rightarrow q$. In problems in which PDFs may significantly differ, this cost will prevent the algorithm be unnecessarily biased by PDFs which are very far away. The exponential kernel has the property of providing a larger weight to smaller distances. The final 2 cost functions, $J_3(A)$ (\ref{e:J3}) and $J_4(A)$ (\ref{e:J4}), operate with the knowledge that the Kullback-Leibler divergence and Hellinger distance are strictly non-increasing given an orthonormal perturbation of the data. Hence, the projection matrix which best preserves the information divergence will be the one that maximizes the information divergence. $J_3(A)$ and $J_4(A)$ adapt $J_1(A)$ and $J_2(A)$, respectively, with this property -- the negative sign on $J_3(A)$ is to coincide with the minimization from (\ref{E:proj_obj1}). The proofs of these properties are presented in Appendices \ref{A:KL_non-increasing} and \ref{A:Hel_non-increasing}, and extending them to the exact Fisher information distance is an area for future work. Note also that $J_2(A)$ and $J_4(A)$ are bounded in the range $[0,N^2]$ due to the bounding of the negative exponential from $[0,1]$. We note that a tighter bound on $J_2(A)$ and $J_4(A)$ is $[0,N(N-1)]$ since $D_{ii}=0$, but as $N\rightarrow\infty$ the difference becomes negligible.

While the choice of cost function is dependent on the problem, the overall projection method ensures that the similarity between data sets is maximally preserved in the desired low-dimensional space, allowing for comparative learning between sets.

\subsection{Gradient Descent}
Gradient descent (or the method of \emph{steepest} descent) allows for the solution of convex optimization problems by traversing a surface or curve in the direction of greatest change, iterating until the minimum is reached. Specifically, let $J(x)$ be a real-valued objective function which is differentiable about some point $x_i$. The direction in which $J(x)$ decreases the fastest, from the point $x_i$, is that of the negative gradient of $J$ at $x_i$, $-\frac{\partial}{\partial x}J(x_i)$. By calculating the location of the next iteration point as
\[
x_{i+1}=x_i-\mu\frac{\partial}{\partial x}J(x_i)
,\]
where $\mu$ is a small number regulating the step size, we ensure that $J(x_i)\geq J(x_{i+1})$. Continued iterations will result in $J(x)$ converging to a local minimum. Gradient descent does not guarantee that the process will converge to an absolute minimum, so typically it is important to initialize $x_0$ near the estimated minimum.

Let $J(A)$ be our objective function, measuring the error between our projected subspace and the full-dimensional space. The direction of the gradient is solved by taking the partial derivative of $J$ w.r.t.~a projection matrix $A$:
\begin{eqnarray}
\frac{\partial}{\partial A}J_1 & = & \sum_i\sum_j \frac{\partial}{\partial A} \left[D_{ij}^2(\mX;A)-2D_{ij}(\mX)D_{ij}(\mX;A) \right] \nonumber \\
& = & \sum_i\sum_j 2\left(D_{ij}(\mX;A)-D_{ij}(\mX) \right)\frac{\partial}{\partial A} D_{ij}(\mX;A) \nonumber \\
\frac{\partial}{\partial A}J_2 & = & \sum_i\sum_j \frac{\partial}{\partial A} \left[{\rm e}^{\left(-2D_{ij}(\mX,A)/c\right)}-2{\rm e}^{-\left(D_{ij}(\mX)+D_{ij}(\mX;A)\right)/c}\right] \nonumber \\
& = & \sum_i\sum_j \frac{2}{c} \left({\rm e}^{-D_{ij}(\mX)/c}-{\rm e}^{-D_{ij}(\mX;A)/c}\right)\, {\rm e}^{-D_{ij}(\mX;A)/c}\, \frac{\partial}{\partial A} D_{ij}(\mX;A) \nonumber \\
\frac{\partial}{\partial A}J_3 & = & -\sum_i\sum_j 2D_{ij}(\mX;A)\frac{\partial}{\partial A} D_{ij}(\mX;A) \nonumber \\
\frac{\partial}{\partial A}J_4 & = & -\sum_i\sum_j \frac{2}{c}\, {\rm e}^{-2D_{ij}(\mX;A)/c}\, \frac{\partial}{\partial A} D_{ij}(\mX;A) \nonumber
.
\end{eqnarray}

Given the direction of the gradient, the projection matrix can be updated as
\Eq{ \label{E:proj_update}
A=A-\mu\frac{\partial}{\partial A}\tilde{J}(A)
,}
where \[\frac{\partial}{\partial A}\tilde{J}=\frac{\partial}{\partial A}J-\frac{1}{2}\left(\left(\frac{\partial}{\partial A}J\right) A^T+A\left(\frac{\partial}{\partial A}J\right)^T\right)A\] is the direction of the gradient, constrained to force $A$ to remain orthonormal (the derivation of this constraint can be found in Appendix \ref{A:Orth}). This process is iterated until the error $J$ converges.

\subsection{IPCA Algorithm}
\begin{algorithm}[t]
\caption{Information Preserving Component Analysis}
\label{a:process}
    \begin{algorithmic}[1]
        \REQUIRE Collection of data sets $\mX=\{\vX_1,\ldots,\vX_N\}$, each of dimension $d$; the desired projection dimension $m$; search step size $\mu$; threshold $\epsilon$
        \STATE Calculate $D(\mX)$, the Fisher information distance matrix
        \STATE Initialize $A_1\in\Real^{m\times d}$ as some orthonormal projection matrix
        \STATE Calculate $D(\mX;A_1)$, the Fisher information distance matrix in the projected space
        \WHILE{$|J_i-J_{i-1}|>\epsilon$}
            \STATE Calculate $\frac{\partial}{\partial A_i}\tilde{J}$, the direction of the gradient, constrained to $AA^T=I$
            \STATE $A_{i+1}=A_i-\mu \frac{\partial}{\partial A_i}\tilde{J}$
            \STATE Calculate $D(\mX;A_{i+1})$
            \STATE Calculate $J$
            \STATE $i=i+1$
        \ENDWHILE
        \ENSURE IPCA Projection $A\in\Real^{m\times d}$
    \end{algorithmic}
\end{algorithm}

The full method for IPCA is described in Algorithm \ref{a:process}. We note that $A_1$ is often initialized as a random orthonormal projection matrix as to not bias the estimation, but this carries the risk of converging to a local minimum. For certain applications it may be beneficial to initialize near some estimated global minimum if that information is available. At this point we stress that we utilize gradient descent due to its ease of implementation. There are more efficient methods of optimization, but that is out of the scope of the current contribution and is an area for future work.

\subsection{Variable Selection}
IPCA may be used as a form of variable selection, as the loading vectors in the linear projection matrix $A$ will be appropriately weighted towards the dimensions which best preserve the information distance between sets within the collection. For example, if two multivariate PDFs $p$ and $q$ are independent and identically distributed in a certain dimension, that dimension will offer zero contribution to the information distance between $p$ and $q$. As such, the information distance is entirely defined by those areas of input space in which $p$ and $q$ differ. When finding a projection which preserves the information distance between $p$ and $q$, $A$ is going to be highly weighted towards the variables which contribute most to that distance. Hence, the loading vectors of $A$ essentially give a ranking of the discriminative value of each variable. This form of variable selection is useful in exploratory data analysis.

\section{Implementation}
\label{S:Details}
We now detail the calculation of the approximation of the Fisher information distance between two realizations of PDFs. Specifically, let $\vX_f$ and $\vX_g$ be realizations of PDFs $f(x)$ and $g(x)$ respectively. Our goal is to calculate both an approximation of the Fisher information distance between the PDFs as well as the direction of the gradient with respect to a projection matrix $A$. Let us now illustrate the difficulties with these computations.

Recall that the Hellinger distance (squared) is defined as
\Eq{\label{e:hellinger}
D_H^2(f(x),g(x))=\int{\left(\sqrt{f(x)}-\sqrt{g(x)}\right)^2dx}
.}
Given the limited (and often unknown) support of $x$ in both $f(x)$ and $g(x)$, it is appropriate to reformat this definition in terms of an expected value with respect to a single density $f(x)$ or $g(x)$:
\Eq{\label{e:e_hellinger}
D_H^2(f(x),g(x))=\left\{\begin{array}{c}
                   \int{\left(1-\sqrt{\frac{g(x)}{f(x)}}\right)^2f(x)dx} \\
                   \int{\left(1-\sqrt{\frac{f(x)}{g(x)}}\right)^2g(x)dx}
                 \end{array}\right.
.}
These equations may be numerically approximated as follows:
\[
\hat{D}_H^2(f(x),g(x))=\left\{\begin{array}{c}
    \frac{1}{n_f}\sum_{i=1}^{n_f}{\left(1-\sqrt{\frac{\hat{g}\left(x_i^{(f)}\right)}{\hat{f}\left(x_i^{(f)}\right)}}\right)^2} \\
    \frac{1}{n_g}\sum_{i=1}^{n_g}{\left(1-\sqrt{\frac{\hat{f}\left(x_i^{(g)}\right)}{\hat{g}\left(x_i^{(g)}\right)}}\right)^2}
\end{array} \right.
,\]
in which $n_f$ and $n_g$ are the number of samples $x_i^{(f)}\in\vX_f$ and $x_i^{(g)}\in\vX_g$, and $\hat{f}(x)$ and $\hat{g}(x)$ are the kernel density estimates of PDFs $f(x)$ and $g(x)$ (see Section \ref{ss:matlab} and Appendix \ref{A:KDE}). The problem with these approximations is that they yield a non-symmetric estimate of the Hellinger distance $\hat{D}_H^2(f(x),g(x))\neq \hat{D}_H^2(g(x),f(x))$. Additionally, the estimate is unbounded from above. By definition the Hellinger distance should be symmetric and bounded by 2 (for the squared distance).

When approximating the Kullback-Leibler divergence, a similar approach of formatting as an expectation may seem natural. The definition of the KL divergence
\Eq{\label{e:kl}
KL(f(x)\|g(x))=\int{f(x)\log \frac{f(x)}{g(x)}\, dx }
}
in turn would be approximated as
\[
\hat{KL}(f\|g)=\frac{1}{n_f}\sum_{i=1}^{n_f}{\log{ \frac{\hat{f}\left(x_i^{(f)}\right)}{\hat{g}\left(x_i^{(f)}\right)}} }
.\]
Note that by definition the KL divergence is not necessarily symmetric, however it is strictly non-negative. This numerical approximation does not guarantee this non-negativity.

\subsection{Metric Calculation}
\label{SS:NumAppr}
We now detail our approximations which do not suffer from the aforementioned pitfalls. Define \[T(x)=\frac{f(x)}{f(x)+g(x)}\] and
\Eq{\label{e:That}
\hat{T}(x)=\frac{\hat{f}(x)}{\hat{f}(x)+\hat{g}(x)}.}
Note that $0\leq T(x) \leq 1$.

For simplicity, let us write $f=f(x)$, $g=g(x)$, and $T=T(x)$. The Hellinger distance (squared) may be computed as follows:
\begin{eqnarray}
D_H^2(f,g) & = & \int{\left(\sqrt{f}-\sqrt{g}\right)^2dx} \nonumber \\
& = & \int{\left(\sqrt{\frac{f}{f+g}}-\sqrt{\frac{g}{f+g}}\right)^2 (f+g)dx} \nonumber \\
& = & \int{\left(\sqrt{T}-\sqrt{1-T}\right)^2 f\, dx } + \int{\left(\sqrt{T}-\sqrt{1-T}\right)^2 g \,dx } \label{e:hellT}.
\end{eqnarray}
Hence, we may now define our numerical approximation of the squared Hellinger distance as:
\begin{eqnarray}
\hat{D}_H^2(f,g) & = & \frac{1}{n_f}\sum_{i=1}^{n_f}{\left(\sqrt{\hat{T}\left(x_i^{(f)}\right)} - \sqrt{1-\hat{T}\left(x_i^{(f)}\right)}\right)^2 } \nonumber \\
& & + \frac{1}{n_g}\sum_{i=1}^{n_g}{\left(\sqrt{\hat{T}\left(x_i^{(g)}\right)} - \sqrt{1-\hat{T}\left(x_i^{(g)}\right)}\right)^2 } ,
\label{e:hellThat}
\end{eqnarray}
which is both symmetric and bounded above by 2.

The same formulation may be implemented when approximating the Kullback-Leibler divergence. Specifically,
\begin{eqnarray}
KL(f\|g) & = & \int{f\log{\frac{f}{g}}\,dx} \nonumber \\
& = & \int{\frac{f}{f+g}\log{\left(\frac{f}{f+g}/\frac{g}{f+g}\right)}(f+g)\,dx} \nonumber \\
& = & \int{T\log{\frac{T}{1-T}f\,dx} } + \int{T\log{\frac{T}{1-T}}g\,dx}. \label{e:klT}
\end{eqnarray}
Hence
\begin{eqnarray}
\hat{KL}(f\|g) & = & \frac{1}{n_f}\sum_{i=1}^{n_f}{\hat{T}\left(x_i^{(f)}\right) \log{\frac{\hat{T}\left(x_i^{(f)}\right)}{1-\hat{T}\left(x_i^{(f)}\right)} } } \nonumber \\
& & + \frac{1}{n_g}\sum_{i=1}^{n_g}{\hat{T}\left(x_i^{(g)}\right) \log{\frac{\hat{T}\left(x_i^{(g)}\right)}{1-\hat{T}\left(x_i^{(g)}\right)} } } ,
\label{e:klThat}
\end{eqnarray}
which no longer suffers from the issue of potential negativity. This value is still non-symmetric, and when dealing with metric learning it is often desirable to have a symmetric dissimilarity metric. Hence, we implement the symmetric KL divergence (\ref{Equation:DKL}) which maybe calculated in a similar manner:
\begin{eqnarray}
D_{KL}(f,g) & = & KL(f\|g)+KL(g\|f) \nonumber \\
& = & \int{f\log{\frac{f}{g}}\,dx} + \int{g\log{\frac{g}{f}}\,dx} \nonumber \\
& = & \int{(f-g)\log{\frac{f}{g}}\,dx} \nonumber \\
& = & \int{(2T-1)\log{\frac{T}{1-T}f\,dx} } + \int{(2T-1)\log{\frac{T}{1-T}}g\,dx}, \label{e:klsymT}
\end{eqnarray}
yielding
\begin{eqnarray}
\hat{D}_{KL}(f,g) & = & \frac{1}{n_f}\sum_{i=1}^{n_f}{(2\hat{T}\left(x_i^{(f)}\right)-1) \log{\frac{\hat{T}\left(x_i^{(f)}\right)}{1-\hat{T}\left(x_i^{(f)}\right)} } } \nonumber \\
& & + \frac{1}{n_g}\sum_{i=1}^{n_g}{(2\hat{T}\left(x_i^{(g)}\right)-1) \log{\frac{\hat{T}\left(x_i^{(g)}\right)}{1-\hat{T}\left(x_i^{(g)}\right)} } } .
\label{e:klsymThat}
\end{eqnarray}

Although not previously discussed, we now illustrate another probabilistic distance measure, the Bhattacharya distance. This measure is a special case of the Chernoff distance,
\[D_{CH}=-\log\int{f(x)^sg(x)^{1-s}}\,dx ,\]
with $s=\frac{1}{2}$. These measure has been utilized for an upper bound on probability of error in classification problems regarding dimensionality reduction \cite{Hsieh:PAMI06}, and is therefore useful in comparative analysis between differing methods. The Bhattacharya distance may be numerically formulated in the same manner as the KL divergence and Hellinger distance:
\begin{eqnarray}
D_B(f,g) & = & -\log\int{\sqrt{f}\sqrt{g}\,dx} \nonumber \\
& = & -\log\int{\sqrt{f}\sqrt{g}\,\frac{f+g}{f+g}\,dx} \nonumber \\
& = & -\log{\left[\int{\sqrt{T(1-T)}f\,dx} + \int{\sqrt{T(1-T)}g\,dx}\right]}  \nonumber \\
\hat{D}_B(f,g) & = & -\log\left[\frac{1}{n_f}\sum_{i=1}^{n_f}{
\sqrt{\hat{T}\left(x_i^{(f)}\right)\left(1-\hat{T}\left(x_i^{(f)}\right)\right)}}\right. \nonumber \\
& & \left.- \frac{1}{n_g}\sum_{i=1}^{n_g}{
\sqrt{\hat{T}\left(x_i^{(g)}\right)\left(1-\hat{T}\left(x_i^{(g)}\right)\right)}}\right]
.
\label{e:bhattaThat}
\end{eqnarray}
Note that the Bharracharya distance is directly related to the Hellinger distance by the transformation $D_B(f,g)=-\log{\left(1-\frac{1}{2}D_H(f,g)^2\right)}$.

In order to numerically approximate these information divergences (\ref{e:hellThat},\ref{e:klThat},\ref{e:klsymThat},\ref{e:bhattaThat}), we calculate $\hat{T}(x)$ as described in Section \ref{ss:matlab}.

\subsection{Gradient Calculation}
\label{SS:Gradient}
In Section \ref{SS:NumAppr} we detailed expressions of the form
\[D=\frac{1}{n_f}\sum_{i=1}^{n_f}G\left(\hat{T}(x_i^{(f)})\right) + \frac{1}{n_g}\sum_{i=1}^{n_g}G\left(\hat{T}(x_i^{(g)})\right),\]
which were used to numerically approximate the Hellinger distance and Kullback-Leibler divergence between PDFs $f(x)$ and $g(x)$. For simplicity, we write
\[D=\frac{1}{n_f}\sum_{i=1}^{n_f}G(\hat{T})\mid_{x_i^{(f)}} + \frac{1}{n_g}\sum_{i=1}^{n_g}G(\hat{T})\mid_{x_i^{(g)}}.\]
Note that we do not continue with the Bhattacharya distance as we are mainly interested in that measure for final comparison to other dimension reduction methods, and we do not calculate the gradient w.r.t.~this distance. If desired, the gradient may be calculated by a analytic transformation of the Hellinger distance gradient.

The gradient of $D$ w.r.t.~some parameter $\theta$, to which $T$ yields some dependency, is defined as
\Eq{\label{e:dgrad}
\frac{\partial D}{\partial \theta} = \frac{1}{n_f}\sum_{i=1}^{n_f}{\frac{\partial G}{\partial T}\frac{\partial T}{\partial \theta}\mid_{x_i^{(f)}} } + \frac{1}{n_g}\sum_{i=1}^{n_g}{\frac{\partial G}{\partial T}\frac{\partial T}{\partial \theta}\mid_{x_i^{(g)}} } ,}
where
\Eq{\label{e:tgrad}
\frac{\partial T}{\partial \theta} = T(1-T)\left( \frac{\partial}{\partial \theta}\log{f} - \frac{\partial}{\partial \theta}\log{g} \right)
.}
This derivation is explained in Appendix \ref{A:Tgrad_derivation}. Substituting (\ref{e:tgrad}) into (\ref{e:dgrad}), the gradient may be numerically approximated as
\begin{eqnarray}
\frac{\partial \hat{D}}{\partial \theta} & = & \frac{1}{n_f}\sum_{i=1}^{n_f}{\hat{T}(1-\hat{T})\frac{\partial G}{\partial T} \left( \frac{\partial}{\partial \theta}\log{f} - \frac{\partial}{\partial \theta}\log{g} \right) \mid_{x_i^{(f)}} } \nonumber \\
& & + \frac{1}{n_g}\sum_{i=1}^{n_g}{\hat{T}(1-\hat{T}) \frac{\partial G}{\partial T} \left( \frac{\partial}{\partial \theta}\log{f} - \frac{\partial}{\partial \theta}\log{g} \right) \mid_{x_i^{(g)}} }
.
\end{eqnarray}

Given this general setting, it is important to recognize that the only difference between the formulation of the Hellinger distance, KL divergence, and symmetric KL divergence is the definition of $G(T)$. Hence, the formulation of the gradient is unchanged for all metrics, given a different definition of $G(T)$. We now derive the value of $T(1-T)\frac{\partial G}{\partial T}$ for each metric.

\subsubsection{Hellinger Distance}

From (\ref{e:hellT}) we see that $G(T)=\left(\sqrt{T}-\sqrt{1-T}\right)^2$. Therefore,
\begin{eqnarray}
\frac{\partial G}{\partial T} & = & (\sqrt{T}-\sqrt{1-T})\left(\frac{1}{\sqrt{T}} + \frac{1}{\sqrt{1-T}} \right) \nonumber \\
& = & \sqrt{\frac{T}{1-T}}-\sqrt{\frac{1-T}{T}} \nonumber \\
& = & \frac{2T-1}{\sqrt{(1-T)T}} \nonumber .
\end{eqnarray}
Hence,
\Eq{\label{e:hell_dgdt}
T(1-T)\frac{\partial G}{\partial T} = \sqrt{T(1-T)} \, (2T-1)
.}

\subsubsection{Kullback-Leibler Divergence}

From (\ref{e:klT}) we see that $G(T)=T\log{\frac{T}{1-T}}$. Therefore,
\begin{eqnarray}
\frac{\partial G}{\partial T} & = & \log{\left(\frac{T}{1-T}\right)}+\frac{1}{1-T} .
\end{eqnarray}
Hence,
\Eq{\label{e:kl_dgdt}
T(1-T)\frac{\partial G}{\partial T} = T(1-T)\log{\left(\frac{T}{1-T}\right)}+T
.}

For the symmetric KL divergence, (\ref{e:klsymT}) yields that $G(T)=(2T-1)\log{\frac{T}{1-T}}$. Therefore,
\[
\frac{\partial G}{\partial T} = 2\log{\left(\frac{T}{1-T}\right)}+\frac{2T-1}{T(1-T)} .\]
Hence,
\Eq{\label{e:klsym_dgdt}
T(1-T)\frac{\partial G}{\partial T} = 2T(1-T)\log{\left(\frac{T}{1-T}\right)}+2T-1
.}

\subsection{Matrix Gradient}
We now specify our abstraction to the specific task for IPCA, which is calculating the gradient of $D$ w.r.t.~the projection matrix $A$. First, let us derive $\frac{\partial}{\partial \theta}\log{f}=\frac{\partial}{\partial A}\log{f(Ax)}$ in which $\hat{f}(Ax)$ may be estimated with kernel density estimation methods described in Appendix \ref{A:KDE}, with kernel locations $x_j^{(f)}\in\vX_f$.
\begin{eqnarray}
\frac{\partial}{\partial A}\log{\hat{f}(Ax)}|_{x^{(f)}} & = & \frac{\partial}{\partial A} \log{\left( \frac{1}{n_f}\sum_{j=1}^{n_f}{\frac{1}{\sqrt{|2\pi H_f|}}\, {\rm e}^{-\frac{1}{2} (x-x_j^{(f)})^T A^T H_f^{-1} A (x-x_j^{(f)}))}}\right)} \nonumber \\
& = & \sum_{j=1}^{n_f}{\bar{W}_{j}^{(f)} \left(-H_f^{-1}A(x-x_j^{(f)})(x-x_j^{(f)})^T\right)} \nonumber \\
& = & -H_f^{-1} A C^{(f)}(x) ,
\end{eqnarray}
where $H_f$ is the kernel bandwidth for set $\vX_f$,
\[
\bar{W}_{j}^{(f)}=\frac{\exp{\left(-\frac{1}{2} (x-x_j^{(f)})^T A^T H_f^{-1} A (x-x_j^{(f)})\right)}}{\sum_{l=1}^{n_f}{\exp{\left(-\frac{1}{2} (x-x_l^{(f)})^T A^T H_f^{-1} A (x-x_l^{(f)})\right)}}} ,
\]
and
\[
C^{(f)}(x)=\sum_{j=1}^{n_f}{\bar{W}_{j}^{(f)} (x-x_j^{(f)})(x-x_j^{(f)})^T}
\]
is the weighted sample covariance around $x$. In the same manner,
\[
\frac{\partial}{\partial A} \log{\left(\hat{g}(Ax)\right)}|_{x^{(g)}}=-H_g^{-1} A C^{(g)}(x),\]
by evaluating the KDE with points $x_j^{(g)}\in\vX_g$.

Finally, we may now define the gradient $\frac{\partial}{\partial A} \hat{D}$ as follows
\begin{eqnarray}
\frac{\partial}{\partial A} \hat{D} & = & \frac{1}{n_f}\sum_{i=1}^{n_f}\hat{T}(x_i^{(f)})(1-\hat{T}(x_i^{(f)}))\frac{\partial G}{\partial T}(x_i^{(f)}) \left[ (-H_f^{-1} A C^{(f)}(x_i^{(f)}))-\ldots\right. \nonumber \\
& & \left.-(-H_g^{-1} A C^{(g)}(x_i^{(f)}))\right] \nonumber \\
& & + \frac{1}{n_g}\sum_{i=1}^{n_g}\hat{T}(x_i^{(g)})(1-\hat{T}(x_i^{(g)}))\frac{\partial G}{\partial T}(x_i^{(g)}) \left[ (-H_f^{-1} A C^{(f)}(x_i^{(g)}))-\ldots \right. \nonumber \\
& & \left.- (-H_g^{-1} A C^{(g)}(x_i^{(g)}))\right] \label{e:grad_final}.
\end{eqnarray}

\subsection{Numerical Implementation}
\label{ss:matlab}
\subsubsection{PDFs and $T(x)$}
To estimate the PDFs $f(x)$ and $g(x)$, let us begin with the following definitions:
\begin{eqnarray}
D_{ij}^{(f,g)} & = & (x_i^{(f)}-x_j^{(g)})^TH_f^{-1} (x_i^{(f)}-x_j^{(g)}) \nonumber \\
W_{ij}^{(f,g)} & = & {\rm e}^{-D_{ij}^{(f,g)}/2} \nonumber \\
W^{(f,g)} & = & \left[W_{ij}^{(f,g)}\right] , \label{e:matlab_defs1}
\end{eqnarray}
which in conjunction with the Gaussian KDE illustrated in Appendix \ref{A:KDE}, yield the density estimates:
\begin{eqnarray}
\hat{f}(x)|_{x^{(\star)}}=\frac{1}{n_\star} \frac{1}{\sqrt{|2\pi H_f|}} W^{(f,\star)}\Vc{1} \nonumber \\
\hat{g}(x)|_{x^{(\star)}}=\frac{1}{n_\star} \frac{1}{\sqrt{|2\pi H_g|}} W^{(g,\star)}\Vc{1}, \label{e:matlab_pdfs}
\end{eqnarray}
where $\Vc{1}$ is the vector of all ones and $\star$ is either $f$ or $g$. Essentially, given the limited support of each $\vX_f$ and $\vX_g$, we approximate the densities and their derivatives w.r.t.~the samples in an appropriate set. Hence, $\hat{f}(x)|_{x^{(\star)}}$ is an $n_\star$ element vector with elements equal to $\hat{f}(x^{(\star)})$, $x^{(\star)}\in\vX_\star$. A similar interpretation holds for $\hat{g}(x)|_{x^{(\star)}}$.

Plugging the density estimates (\ref{e:matlab_pdfs}) into (\ref{e:That}), we calculate our final estimates of $T(x)$:
\begin{eqnarray}
\hat{T}(x^{(f)}) = \frac{1}{\sqrt{|2\pi H_f|}}W^{(f,f)}\Vc{1} ./ \left(\frac{1}{\sqrt{|2\pi H_f|}}W^{(f,f)}\Vc{1}+ \frac{1}{\sqrt{|2\pi H_g|}}W^{(g,f)}\Vc{1}\right) \nonumber \\
\hat{T}(x^{(g)}) = \frac{1}{\sqrt{|2\pi H_f|}}W^{(f,g)}\Vc{1} ./ \left(\frac{1}{\sqrt{|2\pi H_f|}}W^{(f,g)}\Vc{1}+ \frac{1}{\sqrt{|2\pi H_g|}}W^{(g,g)}\Vc{1}\right),
\label{e:matlab_T}
\end{eqnarray}
where the notation $./$ signifies element-by-element vector division.

\subsubsection{Gradient}
We now describe the implementation of (\ref{e:grad_final}). Specifically, let us numerically calculate
\[Z(f,g)=\frac{1}{n_f}\sum_{i=1}^{n_f}{\hat{T}(x_i^{(f)})(1-\hat{T}(x_i^{(f)}))\frac{\partial G}{\partial T}(x_i^{(f)}) \left( H_g^{-1} A C^{(g)}(x_i^{(f)}) \right)}
,\]
and extend towards the 3 other similar formulations such that
\Eq{\label{e:matlab_grad}
\frac{\partial}{\partial A}\hat{D} = Z(f,g)-Z(f,f)+Z(g,g)-Z(g,f)
.}

Let us continue (\ref{e:matlab_defs1}) with the following additional definitions:
\begin{eqnarray}
\bar{W}_{ij}^{(f,g)} & = & W_{ij}^{(f,g)} / (W_{ij}^{(f,g)}\Vc{1}\Vc{1}^T) \nonumber \\
S_{ij}^{(f,g)} & = & \hat{T}(1-\hat{T})\frac{\partial G}{\partial T}|_{x_i^{(f)}}\cdot\bar{W}_{ij}^{(f,g)} \nonumber \\
S^{(f,g)} & = & \left[S_{ij}^{(f,g)}\right] . \label{e:matlab_defs2}
\end{eqnarray}
The formulation continues as follows:
\begin{eqnarray}
Z(f,g) & = & \frac{1}{n_f}H_g^{-1} A \sum_{i=1}^{n_f}{ \sum_{j=1}^{n_g}{ \hat{T}(1-\hat{T})\frac{\partial G}{\partial T}|_{x_i^{(f)}}\cdot\bar{W}_{ij}^{(f,g)} (x_i^{(f)}-x_j^{(g)})(x_i^{(f)}-x_j^{(g)})^T}} \nonumber \\
& = & \frac{1}{n_f}H_g^{-1} A \sum_{i=1}^{n_f}{ \sum_{j=1}^{n_g}{ S_{ij}^{(f,g)} (x_i^{(f)}-x_j^{(g)})(x_i^{(f)}-x_j^{(g)})^T}} \nonumber \\
& = & \frac{1}{n_f}H_g^{-1} A \left[\vX_f {\rm diag}(S^{(f,g)}\Vc{1})\vX_f^T - \vX_f S^{(f,g)} \vX_g^T \right.  \nonumber \\
& & -\left. \vX_g (S^{(f,g)})^T \vX_f^T + \vX_g {\rm diag}((S^{(f,g)})^T\Vc{1})\vX_g^T \right].
\label{e:zfg}
\end{eqnarray}

Equation (\ref{e:zfg}) and similar formulations (replacing $f$ and $g$ where appropriate) may be substituted into (\ref{e:matlab_grad}) to obtain the final calculation of the gradient $\frac{\partial}{\partial A}\hat{D}$.

\section{Conclusions}
\label{S:Conclusions}
The assumption that high-dimensional data lies on a Riemannian manifold in Euclidean space is often based on the ease of implementation due to the wealth of knowledge and methods based on Euclidean space. This assumption is not viable in many problems of practical interest, as there is often no straightforward and meaningful Euclidean representation of the data. In these situations it is more appropriate to assume the data is a realization of some PDF lying on a statistical manifold. Using information geometry, we have shown the ability to find a low-dimensional embedding of the manifold, which allows us to not only find the natural separation of the data, but to also reconstruct the original manifold and visualize it in a low-dimensional Euclidean space. This allows the use of many well known learning techniques which work based on the assumption of Euclidean data.

We have also offered an unsupervised method of dimensionality reduction which preserves the information distances between high-dimensional data sets in the low-dimensional projection space. Information Preserving Component Analysis finds a low-dimensional projection space designed to maintain the similarities between data sets, which enables a comparative analysis in how sets from different generative models occupy the projection space. Additionally, analysis of the loading vectors in the projection matrix allows for a means of variable selection, as the variables which are most crucial to preserving the information distances will have the largest loading values.

We now stress that this framework is not meant to be the seminal result of information geometric dimensionality reduction, but a means for looking at common problems from a new angle. We have kept our formulations intentionally abstract as we make no claims as to which metric, implementation, or cost function is optimal. On the contrary, those choices are determined by the problem of interest. Instead, we have presented a framework to which others may tailor specifically to their needs, and we offer specific methods of implementation which may be immediately used.

\appendix
\subsection{Appendix: Kernel Density Estimation}
\label{A:KDE}
We now illustrate the derivation of the kernel density estimate (KDE) of the PDF $f(x)$ of the realization $\vX_f$. We utilize Gaussian kernels as the quadratic properties will be useful in implementation. Specifically, the KDE of a PDF is defined as
\Eq{ \label{Equation:KDE}
        \hat{f}(x)=\frac{1}{n_f\cdot h}\sum_{j=1}^{n_f}{K\left(\frac{x-x_j}{h}\right)}
    ,}
where $n_f$ is the number of sample points $x_j\in\vX_f$, $K$ is some kernel satisfying the properties
\[
K(x)\ge0, \,\forall x\in\mX
,\]
\[\int K(x)\,dx=1,\]
and $h$ is the bandwidth or smoothing parameter. By utilizing the Gaussian kernel
\[
K(x)=\frac{1}{|2\pi\Sigma|^{1/2}}\exp\left(-\frac{1}{2}x^T\Sigma^{-1}x\right)
,\]
where $\Sigma$ is the covariance of the kernel, we may combine the smoothing parameter vector $h$ with $\Sigma$ (ie.~$\Sigma=I$) such that $H_f={\rm diag}(h)$. Note that we implement a vector bandwidth such that our Gaussian kernels are ellipses rather than spheres. There are a variety of methods for determining this bandwidth parameter; we choose to implement the maximal smoothing principle \cite{Terrell:JASA90}. This yields the a final kernel density estimate of
\Eq{\label{e:kde_gauss}
\hat{f}(x)=\frac{1}{n_f}\sum_{j=1}^{n_f}{\frac{1}{\sqrt{|2\pi H_f|}}\exp\left(-\frac{1}{2}(x-x_j)^T H_f^{-1}(x-x_j)\right)}
.}

Let us now make the following definitions:
\begin{eqnarray}
D_{j}^{(f)} & = & (x-x_j^{(f)})^T H_f^{-1} (x-x_j^{(f)}) \nonumber \\
W^{(f)} & = & {\rm e}^{-D^{(f)}/2} , \label{e:kde_defs}
\end{eqnarray}
where $D_j^{(f)}$ is a Mahalanobis distance between the point $x$ and sample points $x_j^{(f)}\in \vX_f$, and $D^{(f)}$ is the vector with elements $D_j^{(f)}$. Substituting (\ref{e:kde_defs}) into (\ref{e:kde_gauss}), we obtain
\Eq{\label{e:kde_final}
\hat{f}(x)=\frac{1}{n_f}\sum_{j=1}^{n_f}{\frac{1}{\sqrt{|2\pi H_f|}}W^{(f)}}
,}
the KDE approximation of the PDF generating $\vX_f$.

\subsection{Appendix: Proof of strictly non-increasing property of KL divergence w.r.t.~an orthonormal data projection}
\label{A:KL_non-increasing}
Here we prove that the KL divergence between the PDFs of $x$ and $x'$ is greater or equal to the KL divergence between the PDFs of $y=Ax$ and $y'=Ax'$, respectively, where $A$ satisfies $AA^T=I$.
\begin{thm}
Let RVs $X,X' \in \Real^n$ have PDFs $f_X$ and $f_{X'}$, respectively. Using the $d\times n$ matrix $A$ satisfying $A A^T = I_d$, construct RVs $Y,Y' \in
\Real^d$ such that $Y=AX$ and $Y'=AX'$. The following relation holds:
\begin{eqnarray}
KL(f_X\|f_{X'}) \ge KL(f_Y\|f_{Y'}),
\end{eqnarray}
where $f_Y$ and $f_{Y'}$ are the PDFs of $Y,Y'$, respectively.
\end{thm}
The proof is in two parts. First, we show that the KL divergence is constant over an arbitrary dimension preserving orthogonal transformation. Next, we show that the same truncation of two random vectors does not increase KL.

Let $M$ be an $n \times n$ orthonormal matrix, i.e.,  $MM^T = I_n$ and $M^TM = I_n$. Define the random vectors $V,V' \in \Real^n$ as follows $V=MX$ and $V'=MX$. By a change of variables, we have
\begin{eqnarray}\label{eq:v}
f_V& :& f_V(v)  =  f_X (M^T v) \\
f_{V'}&:& f_{V'}(v')  =  f_{X'} (M^T v').
\end{eqnarray}
Note that the Jacobian of the transformation is $1$ and $M^T$ is the inverse of the transformation both due to the orthonormality of $M$.
The KL divergence between $V$ and $V'$ is given by
\begin{eqnarray}\label{eq:klv}
KL(f_V\|f_{V'}) = \int f_V(v) \log \frac{f_V(v)}{f_{V'}(v)} dv.
\end{eqnarray}
Substituting the PDFs from (\ref{eq:v}) into (\ref{eq:klv}), we have
\begin{eqnarray}\label{eq:5}
KL(f_V\|f_{V'}) = \int f_X(M^T v) \log \frac{f_X(M^T v)}{f_{X'}(M^T v)} dv.
\end{eqnarray}
Next, using the orthonormality of $M$ we replace  $x= M^T v$ and $dx =dv$ in (\ref{eq:5}) and obtain
\begin{eqnarray}\label{eq:6}
KL(f_V\|f_{V'}) = \int f_X(x) \log \frac{f_X(x)}{f_{X'}(x)} dx = KL(f_X\|f_{X'})
\end{eqnarray}
and the KL divergence remains the same.

We proceed with the second part of the proof. Consider the random vector $V$ as a concatenation of RVs $Y$ and $Z$: $V^T=[Y^T Z^T]$. If we write matrix $M$ as $M^T=[A^T,B^T]$ where $A$ is $d\times n$ and $B$ is $(n-d)\times n$, then $Y=AX$,  $Y'=AX'$, and $A$ satisfies $AA^T=I_d$ (but not $A^T A = I_n$). Since $V^T=[Y^T Z^T]$, we have $KL(f_{V}\|f_{V'})= KL(f_{YZ}\|f_{Y'Z'})$ and by virtue of (\ref{eq:6})
\begin{eqnarray}\label{eq:7}
KL(f_X\|f_{X'}) =  KL(f_{YZ}\|f_{Y'Z'}).
\end{eqnarray}
Next, we use the following lemma:
\begin{lemma}
Let $Y,Y' \in \Real^d$ and $Z,Z' \in \Real^{n-d}$ be RVs and denote: the joint PDF of $Y$ and $Z$ by $f_{YZ}$, the joint PDF of $Y'$ and $Z'$ by $f_{Y'Z'}$, the marginal PDF of $Y$ by $f_Y$, the marginal PDF of $Y'$ by $f_{Y'}$, the conditional PDF of $Z$ by $f_{Z|Y}$, and the conditional PDF of $Z'$ by $f_{Z'|Y'}$. The following holds:
\begin{eqnarray}\label{kl:eq}
\int f(y) KL ( f_{Z|Y}\|f_{Z'|Y'} ) dy =
KL ( f_{YZ}\|f_{Y'Z'})  -
KL (f_Y\|f_{Y'}).
\end{eqnarray}
\end{lemma}
This may be proven as follows:
\begin{eqnarray}
\int f(y) KL ( f(z|y)\|g(z|y)) dy = \nonumber \\
\int f(y) \int  f(z|y) \log \frac{f(z|y)}{g(z|y)} dz dy  = \nonumber \\
\iint f(y,z)    \log \frac{f(y,z)/f(y)}{g(y,z)/g(y)} dz dy = \nonumber \\
\iint f(y,z)   \left( \log \frac{f(y,z)}{g(y,z)} - \log \frac{f(y)}{g(y)} \right) dz dy = \nonumber \\
\iint f(y,z)    \log \frac{f(y,z)}{g(y,z)}  dz dy - \iint f(y,z)  \log \frac{f(y)}{g(y)} dz dy = \nonumber \\
KL ( f(y,z)\|g(y,z))  - \iint f(y,z)  dz \log \frac{f(y)}{g(y)}  dy = \nonumber \\
KL ( f(y,z)\|g(y,z))  - \int f(y)  \log \frac{f(y)}{g(y)}  dy = \nonumber  \\
KL ( f(y,z)\|g(y,z)) - KL (f(y)\|g(y)).
\end{eqnarray}
Identifying that the LHS of (\ref{kl:eq}) is non-negative, we immediately obtain the following corollary:
\begin{cor}\label{cor:1}
Let $Y,Y' \in \Real^d$ and $Z,Z' \in \Real^{n-d}$ be RVs and denote: the joint PDF of $Y$ and $Z$ by $f_{YZ}$, the joint PDF of $Y'$ and $Z'$ by $f_{Y'Z'}$, the marginal PDF of $Y$ by $f_Y$, and  the marginal PDF of $Y'$ by $f_{Y'}$. The following holds:
\begin{eqnarray}\label{eq:decr:kl}
KL(f_{YZ}\|f_{Y'Z'}) \ge KL(f_{Y}\|f_{Y'}).
\end{eqnarray}
\end{cor}
This corollary suggests that KL must not increase as a result of marginalization. Application of (\ref{eq:decr:kl}) from Corollary \ref{cor:1} to (\ref{eq:7}), yields the desired result
\begin{eqnarray}
KL(f_{X}\|f_{X'}) \ge KL(f_{Y}\|f_{Y'}).
\end{eqnarray}

\subsection{Appendix: Proof of strictly non-increasing property of Hellinger distance w.r.t.~an orthonormal data projection}
\label{A:Hel_non-increasing}
Here we prove that the Hellinger distance between the PDFs of $x$ and $x'$ is greater or equal to the Hellinger distance between the PDFs of $y=Ax$ and $y'=Ax'$, respectively, where $A$ satisfies $AA^T=I$. Note that much of this derivation is repetitive to that of the KL divergence in Appendix \ref{A:KL_non-increasing}, but we include all steps here for completeness.
\begin{thm}
Let RVs $X,X' \in \Real^n$ have PDFs $f_X$ and $f_{X'}$, respectively. Using the $d\times n$ matrix $A$ satisfying $A A^T = I_d$, construct RVs $Y,Y' \in
\Real^d$ such that $Y=AX$ and $Y'=AX'$. The following relation holds:
\begin{eqnarray}
D_H(f_X,f_{X'}) \ge D_H(f_Y,f_{Y'}),
\end{eqnarray}
where $f_Y$ and $f_{Y'}$ are the PDFs of $Y,Y'$, respectively.
\end{thm}
The proof is in two parts. First, we show that the Hellinger distance is constant over an arbitrary dimension preserving orthogonal transformation. Next, we show that the same truncation of two random vectors does not increase the Hellinger distance.

Let $M$ be an $n \times n$ orthonormal matrix, i.e.,  $MM^T = I_n$ and $M^TM = I_n$. Define the random vectors $V,V' \in \Real^n$ as follows $V=MX$ and $V'=MX$. By a change of variables, we have
\begin{eqnarray}\label{eq:hel_v}
f_V& :& f_V(v)  =  f_X (M^T v) \\
f_{V'}&:& f_{V'}(v')  =  f_{X'} (M^T v').
\end{eqnarray}
Note that the Jacobian of the transformation is $1$ and $M^T$ is the inverse of the transformation both due to the orthonormality of $M$. The squared Hellinger distance between $V$ and $V'$ is given by
\begin{eqnarray}\label{eq:hel_klv}
D_H^2(f_V,f_{V'}) = \int{\left(\sqrt{f_V(v)}-\sqrt{f_{V'}(v)}\right)^2 dv}.
\end{eqnarray}
Substituting the PDFs from (\ref{eq:hel_v}) into (\ref{eq:hel_klv}), we have
\begin{eqnarray}\label{eq:hel_5}
D_H^2(f_V,f_{V'}) = \int{\left(\sqrt{f_X(M^Tv)}-\sqrt{f_{X'}(M^Tv)}\right)^2 dv}.
\end{eqnarray}
Next, using the orthonormality of $M$ we replace  $x= M^T v$ and $dx =dv$ in (\ref{eq:5}) and obtain
\begin{eqnarray}\label{eq:hel_6}
D_H^2(f_V,f_{V'}) = \int{\left(\sqrt{f_X(x)}-\sqrt{f_{X'}(x)}\right)^2 dv } = D_H^2(f_X,f_{X'})
\end{eqnarray}
and the squared Hellinger distance remains the same.

We proceed with the second part of the proof. Consider the random vector $V$ as a concatenation of RVs $Y$ and $Z$: $V^T=[Y^T Z^T]$. If we write matrix $M$ as $M^T=[A^T,B^T]$ where $A$ is $d\times n$ and $B$ is $(n-d)\times n$, then $Y=AX$,  $Y'=AX'$, and $A$ satisfies $AA^T=I_d$ (but not $A^T A = I_n$). Since $V^T=[Y^T Z^T]$, we have $D_H^2(f_{V},f_{V'})= D_H^2(f_{YZ},f_{Y'Z'})$ and by virtue of (\ref{eq:hel_6})
\begin{eqnarray}\label{eq:hel_7}
D_H^2(f_X,f_{X'}) =  D_H^2(f_{YZ},f_{Y'Z'}).
\end{eqnarray}
Next, we use the following lemma:
\begin{lemma}
Let $Y,Y' \in \Real^d$ and $Z,Z' \in \Real^{n-d}$ be RVs and denote: the joint PDF of $Y$ and $Z$ by $f_{YZ}$, the joint PDF of $Y'$ and $Z'$ by $f_{Y'Z'}$, the marginal PDF of $Y$ by $f_Y$, the marginal PDF of $Y'$ by $f_{Y'}$, the conditional PDF of $Z$ by $f_{Z|Y}$, and the conditional PDF of $Z'$ by $f_{Z'|Y'}$. The following holds:
\begin{eqnarray}\label{kl:hel_eq}
D_H^2(f_{YZ},f_{Y'Z'}) - D_H^2(f_{Y},f_{Y'}) \geq 0.
\end{eqnarray}
\end{lemma}
The proof this Lemma begins as follows:
\begin{eqnarray}
D_H^2(f(y,z),g(y,z)) - D_H^2(f(y),g(y)) = \nonumber \\
\int{\int{\left(\sqrt{f(y,z)}-\sqrt{g(y,z)}\right)^2 dy}dz} - \int{\left(\sqrt{f(y)}-\sqrt{g(y)}\right)^2 dy} = \nonumber \\
\int{\int{f(y,z)+g(y,z)-2\sqrt{f(y,z)g(y,z)} dy}dz} - \int{f(y)+g(y)-2\sqrt{f(y)g(y)} dy} = \nonumber \\
-2\int{\int{\sqrt{f(y,z)g(y,z)} dy}dz} + 2\int{\sqrt{f(y)g(y)} dy} = \nonumber \\
2\left[\int{\sqrt{f(y)g(y)} dy} - \int{\int{\sqrt{f(y,z)g(y,z)} dy}dz}\right]. \nonumber
\end{eqnarray}
We may now continue by showing $\int{\int{\sqrt{f(y,z)g(y,z)}\, dy}\,dz} \le \int{\sqrt{f(y)g(y)}\, dy}$:
\begin{eqnarray}
\int{\int{\sqrt{f(y,z)g(y,z)}\, dy}\,dz} & = & \int{\int{\sqrt{f(y)f(z|y)g(y)g(z|y)}\, dy}\,dz} \label{e:bayes} \\
& = & \int{\int{\sqrt{f(y)g(y)}\sqrt{f(z|y)g(z|y)}\, dy}\,dz} \nonumber \\
& = & \int{\sqrt{f(y)g(y)}\left(\int{\sqrt{f(z|y)g(z|y)} dz}\right)dy} \nonumber \\
& \leq & \int{\sqrt{f(y)g(y)}\left(\int{\sqrt{f(z|y)}^2 dz}\right)^{\frac{1}{2}} \left(\int{\sqrt{g(z|y)}^2 dz}\right)^{\frac{1}{2}} dy} \nonumber \\ \label{e:cauchy} \\
& = & \int{\sqrt{f(y)g(y)}\left(\int{f(z|y) dz}\right)^{\frac{1}{2}} \left(\int{g(z|y) dz}\right)^{\frac{1}{2}} dy} \nonumber \\
& = & \int{\sqrt{f(y)g(y)}\left(1\right)^{\frac{1}{2}} \left(1\right)^{\frac{1}{2}} dy} \nonumber \\
& = & \int{\sqrt{f(y)g(y)}\, dy}. \nonumber
\end{eqnarray}
Note that (\ref{e:bayes}) used Bayes rule and (\ref{e:cauchy}) used the Cauchy-Schwartz inequality. We now immediately obtain the following corollary:
\begin{cor}\label{cor:1}
Let $Y,Y' \in \Real^d$ and $Z,Z' \in \Real^{n-d}$ be RVs and denote: the joint PDF of $Y$ and $Z$ by $f_{YZ}$, the joint PDF of $Y'$ and $Z'$ by $f_{Y'Z'}$, the marginal PDF of $Y$ by $f_Y$, and  the marginal PDF of $Y'$ by $f_{Y'}$. The following holds:
\begin{eqnarray}\label{eq:decr:kl}
D_H^2(f_{YZ},f_{Y'Z'}) \ge D_H^2(f_{Y},f_{Y'}).
\end{eqnarray}
\end{cor}
This corollary suggests that the squared Hellinger distance must not increase as a result of marginalization. Without loss of generality, due to the monotonic behavior of the square root function, the same may be said for the strict Hellinger distance, yielding the desired result
\begin{eqnarray}
D_H(f_{X},f_{X'}) \ge D_H(f_{Y},f_{Y'}).
\end{eqnarray}

\subsection{Appendix: Orthonormality Constraint on Gradient Descent}
\label{A:Orth}
We derive the orthonormality constraint for our gradient descent optimization in the following manner; solving
\[A=\arg\min_{A:AA^T=I} J(A)
,\]
where $I$ is the identity matrix. Using Lagrangian multiplier $M$, this is equivalent to solving
\[A=\arg\min_A \tilde{J}(A)
,\]
where $\tilde{J}(A)=J(A)+{\rm tr}(A^TMA)$. We can iterate the projection matrix A, using gradient descent, as:
\Eq{\label{e:a_update}
A_{i+1}=A_i-\mu\frac{\partial}{\partial A}\tilde{J}(A_i)
,}
where $\frac{\partial}{\partial A}\tilde{J}(A)=\frac{\partial}{\partial A} J(A)+(M+M^T)A$ is the gradient of the cost function w.r.t. matrix $A$. To ease notation, let $\Delta \triangleq \frac{\partial}{\partial A} J(A_i)$ and $\tilde{\Delta} \triangleq \frac{\partial}{\partial A} \tilde{J}(A_i)$. Continuing with the constraint $A_{i+1}A_{i+1}^T=I$, we right-multiply (\ref{e:a_update}) by $A_{i+1}^T$ and obtain
\[
0=-\mu A_i\tilde{\Delta}^T-\mu\tilde{\Delta} A_i^T+\mu^2\tilde{\Delta}\tilde{\Delta}^T
,\]
\Eq{\label{e:derive1}
\mu\tilde{\Delta}\tilde{\Delta}^T=\tilde{\Delta}A^T+A \tilde{\Delta}^T
,}
\[
\mu(\Delta + (M + M^T)A)(\Delta + (M+M^T)A)^T = (\Delta A(M+M^T)A)A^T + A(\Delta A^T(M+M^T)A)
.\]
Let $Q=M+M^T$, hence $\tilde{\Delta}=\Delta+QA$. Substituting this into (\ref{e:derive1}) we obtain:
\[
\mu(\Delta \Delta^T+QA\Delta^T + \Delta A^TQ + QQ^T) = \Delta A^T + A\Delta^T + 2Q
.\]
Next we use the Taylor series expansion of $Q$ around $\mu=0$: $Q=\sum_{j=0}^\infty\mu^j Q_j$. By equating corresponding powers of $\mu$ (i.e.~$\frac{\partial^j}{\partial\mu^j}|_{\mu=0}=0$), we identify:
\[
Q_0=-\frac{1}{2}(\Delta A^T + A\Delta ^T)
,\]
\[
Q_1=\frac{1}{2}(\Delta + Q_0A)(\Delta + Q_0A)^T
.\]
Replacing the expansion of $Q$ in $\tilde{\Delta}=\Delta + QA$:
\[
\tilde{\Delta}=\Delta-\frac{1}{2}(\Delta A^T+A\Delta^T)A + \mu\, Q_1 A + \mu^2\, Q_2A +\ldots
.\]
Finally, we would like to assure a sufficiently small step size to control the error in forcing the constraint due to a finite Taylor series approximation of $Q$. Using the $L_2$ norm of $\tilde{\Delta}$ allows us to calculate an upper bound on the Taylor series expansion:
\[
\|\tilde{\Delta}\| \, \leq \, \|\Delta-\frac{1}{2}(\Delta A^T+A\Delta^T)A\| + \mu\,\| Q_1 A\| + \mu^2\,\| Q_2A\| +\ldots
.\]
We condition the norm of the first order term in the Taylor series approximation to be significantly smaller than the norm of the zeroth order term. If $\mu \ll \|\Delta-\frac{1}{2}(\Delta A^T+A\Delta^T)A\| / \|Q_1 A\|$ then:
\Eq{ \label{e:final_grad}
\frac{\partial}{\partial A}\tilde{J}(A)=\frac{\partial}{\partial A} J(A)-\frac{1}{2}\left(\left(\frac{\partial}{\partial A} J(A)\right) A^T+A\left(\frac{\partial}{\partial A} J(A)^T\right)\right)A
}
is a good approximation of the gradient constrained to $A A^T=I$. We omit the higher order terms as we experimentally find that they are unnecessary, especially as even $\mu^2\rightarrow 0$. We note that while there are other methods for forcing the gradient to obey orthogonality \cite{Edelman&EtAl:JMAA99,Douglas:ICASSP07}, we find our method is straightforward and sufficient for our purposes.

\subsection{Appendix: Gradient of $T$}
\label{A:Tgrad_derivation}
Calculation of the gradient of $T=\frac{f}{f+g}$ w.r.t.~some parameter $\theta$. Let $F_\theta = \frac{\partial}{\partial \theta}F$ for some arbitrary function $F$.
\begin{eqnarray}
\frac{\partial T}{\partial \theta} & = & T \frac{\partial}{\partial \theta} \log{T} \nonumber \\
& = & T \frac{\partial}{\partial \theta} \left(\log{f}-\log{(f+g)}\right) \nonumber \\
& = & T\left(\frac{f_\theta}{f} - \frac{f_\theta+g_\theta}{f+g}\right) \nonumber \\
& = & T\left(\frac{f_\theta g - g_\theta f}{f(f+g)}\right) \nonumber \\
& = & T\left((1-T)\frac{f_\theta}{f}-\frac{g}{f+g}\frac{g_\theta}{g}\right) \nonumber \\
& = & T\left((1-T)(\log{f})_\theta-(1-T)(\log{g})_\theta\right) \nonumber \\
& = & T(1-T)\left(\frac{\partial}{\partial \theta}\log{f} - \frac{\partial}{\partial \theta}\log{g}\right) \nonumber
\end{eqnarray}

\bibliography{ref}
\bibliographystyle{plain}

\end{document}